%% file: main.tex
\documentclass[letterpaper, 10 pt, journal, twoside]{IEEEtran}
\IEEEoverridecommandlockouts
\input{00-imports}

\input{00-defs}

\newcommand{\orcid}[1]{\href{https://orcid.org/#1}{\includegraphics[width=0.6em]{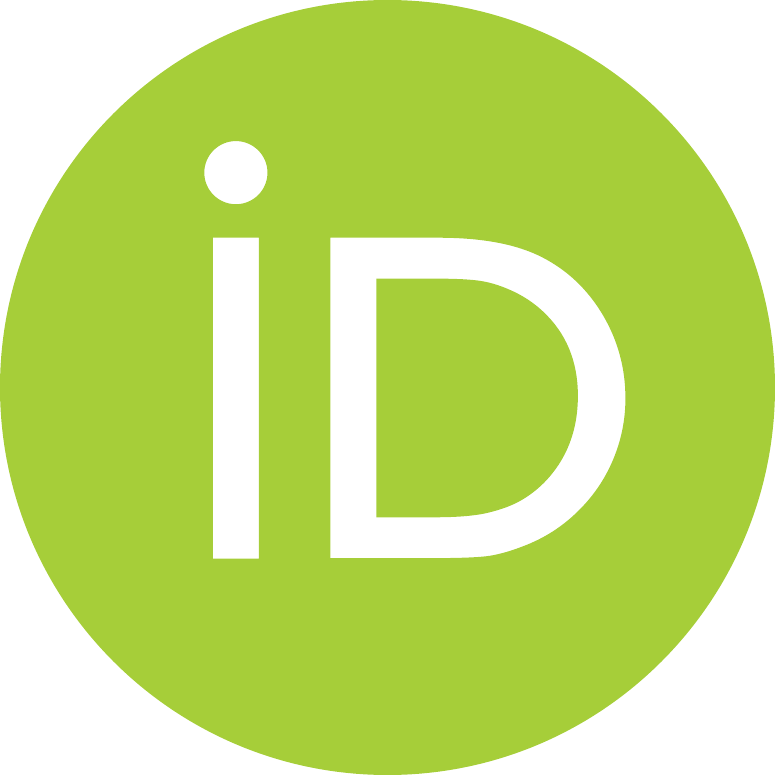}}}
\newcommand{\videoURL}{https://youtu.be/I08UywVVhN4}

\begin{document}
	
	\title{
		RL-Augmented MPC for Non-Gaited 
		Legged and Hybrid Locomotion 
	}
	
	\author{Andrea~Patrizi\orcid{0000-0002-5571-3716},~Carlo~Rizzardo\orcid{0009-0007-0118-5384},~Arturo~Laurenzi\orcid{0000-0002-9065-1266},~Francesco~Ruscelli\orcid{0000-0002-4936-8389},~\\Luca~Rossini\orcid{0000-0002-2114-3823}~and~Nikos~G.~Tsagarakis\orcid{0000-0002-9877-8237}
		\thanks{This work was supported by the European Union's Horizon Europe Framework Programme under grant agreement No 101070596}
		\thanks{All authors are with the Humanoids and Human-Centred Mechatronics (HHCM) lab, Istituto Italiano di Tecnologia (IIT), Via San Quirico 19d, 16163 Genova}
		\thanks{Andrea Patrizi is also with the Department of Informatics, Bioengineering, Robotics and Systems Engineering, University of Genova, Via All’Opera Pia 13, 16145 Genova.}
	}
	
	\maketitle
	
	\begin{abstract}
	We propose a contact-explicit hierarchical architecture coupling Reinforcement Learning (RL) and Model Predictive Control (MPC), where a high-level RL agent provides gait and navigation commands to a low-level locomotion MPC. This offloads the combinatorial burden of contact timing from the MPC by learning acyclic gaits through trial and error in simulation. We show that only a minimal set of rewards and limited tuning are required to obtain effective policies.
	We validate the architecture in simulation across robotic platforms spanning $\mathbf{50}\,$kg to $\mathbf{120}\,$kg and different MPC implementations, observing the emergence of acyclic gaits and timing adaptations in flat-terrain legged and hybrid locomotion, and further demonstrating extensibility to non-flat terrains. Across all platforms, we achieve zero-shot sim-to-sim transfer without domain randomization, and we further demonstrate zero-shot sim-to-real transfer without domain randomization on Centauro, our $\mathbf{120}\,$kg wheeled-legged humanoid robot. We make our software framework and evaluation results publicly available at \url{https://github.com/AndrePatri/AugMPC}.
	\end{abstract}
	
	\begin{IEEEkeywords}
		Whole-Body Motion Planning and Control; Reinforcement Learning; Legged Robots.
	\end{IEEEkeywords}

	\input{01-introduction}
	\input{02-related_works}
	\input{03-overview}
	\input{04-formulation}
	\input{05-results}

	\input{06-future_work}
	\bibliography{IEEEabrv,bibliography/refs}

\end{document}

%% file: 00-imports.tex
\usepackage{amsmath,amsfonts,amssymb,bm, mathtools}
\usepackage{algorithmic}
\usepackage{algorithm}
\usepackage{array}
\usepackage[caption=false,font=normalsize,labelfont=sf,textfont=sf]{subfig}
\usepackage{textcomp}
\usepackage{stfloats}
\usepackage{url}
\usepackage{verbatim}
\usepackage{graphicx}
\usepackage{cite}

\usepackage{url}
\usepackage{multicol}
\usepackage[bookmarks=true]{hyperref}
\usepackage{color}
\usepackage{tabularx}
\usepackage{booktabs}
\usepackage[table]{xcolor}
\usepackage{csquotes}
\usepackage{colortbl}
\usepackage{tikz}

%% file: 00-defs.tex
\newif\ifhighlightchanges
\highlightchangesfalse

\newcommand{\revblock}[1]{\ifhighlightchanges{\color{blue}{#1}}\else#1\fi}

\newcommand{\vc}[1]{\bm#1}

\newcommand{\ximpc}{\vc{\xi}_{\mathrm{MPC}}}
\newcommand{\chimpcj}{{\chi}_{\mathrm{MPC}}^{(j)}}
\newcommand{\chimpc}{\vc{\chi}_{\mathrm{MPC}}}
\newcommand{\feetvrefk}{\vc{v}_{\mathrm{ref},\,z}^{{(j,\,m)}}}
\newcommand{\feetvref}{\vc{v}_{\mathrm{ref},\,z}^{{(j)}}}
\newcommand{\drainedpow}{\max\left(0,\, \left\langle{\dot{\vc{q}}_{\mathrm{jnt}}\cdot{\vc{\tau}}_{\mathrm{jnt}}}\right\rangle\right)}

\definecolor{darkgreen}{rgb}{0.0, 0.5, 0.0}

%% file: 01-introduction.tex
\section{Introduction}
Legged biological systems leverage a combination of sensory feedback, reflexive responses and online re-planning, exhibiting remarkable agility and robustness. In particular, one key skill that natural systems possess is the ability to dynamically choose and adapt gaits based on task demands and sensory information. Interest in replicating these skills on artificial systems has long been a focus of research in legged locomotion~\cite{survey:wensing2023optimization}. \\
The approaches developed over the years classically fall into two main categories: model-based and model-free. Model-based techniques exploit a model of the system dynamics to aid and guide the decision-making process, which is typically formulated as a finite horizon Optimal Control Problem (OCP)~\cite{survey:wensing2023optimization}. The most successful model-based method for locomotion, Model Predictive Control (MPC), has been shown to be capable of remarkable performance over the years~\cite{mpc:tassa2012synthesis,mpc:di2018dynamic,mpc:neunert2018whole,hloc:bjelonic2020rolling,hloc:bjelonic2021whole,mpc:grandia2023perceptive,to:mastalli2023inverse}, thanks to its online re-planning, its interpretability and explicit constraint handling. However, optimizing over contacts leads to a Mixed-Integer Nonlinear Programming problem, which is often too complex to solve online. Therefore, simplified dynamics models and/or heuristics are commonly used to make the problem tractable in real-time~\cite{survey:wensing2023optimization}.\\
Model-free approaches offer a different perspective on the contact timing problem: they forgo explicit dynamics modeling and instead learn policies directly through interaction with the environment~\cite{drl:haarnoja2018soft,drl:agarwal2023legged,drl:rudin2022learning}. Most state-of-the-art model-free Reinforcement Learning (RL) formulations circumvent the combinatorial nature of contact scheduling by training joint-space, contact-implicit policies. However, these approaches often depend heavily on domain randomization and reward shaping~\cite{drl:rudin2022learning,rl:mjx_playground,drl:lee2020learning}, leading to sample inefficiency and increased task-specific tuning.
\begin{figure}[t]
	\includegraphics[width=\columnwidth]{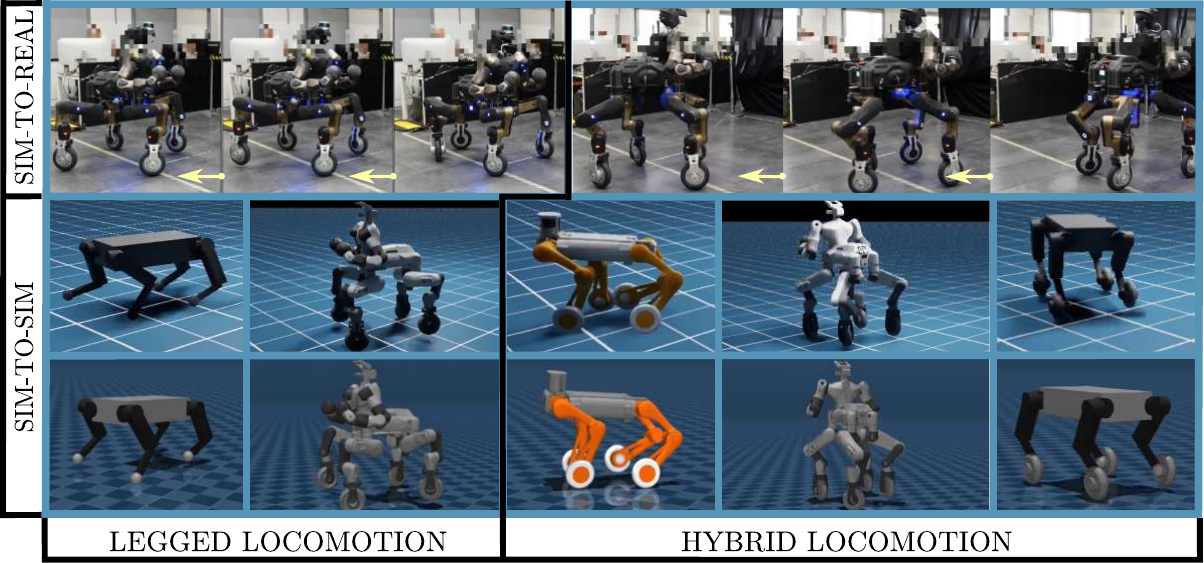}
	\caption{Our hierarchical RL-MPC architecture has been trained and evaluated on several robotic platforms with different morphologies and weight distributions ($50~\mathrm{kg}$ to $120~\mathrm{kg}$), successfully transferred zero-shot across different domains without any domain randomization, and applied to both hybrid and standard locomotion tasks. Supplementary videos are available at \url{\videoURL}.}
	\label{fig:intro}
\end{figure}

We propose a hierarchical architecture that couples a high-level RL policy with a low-level MPC.
Unlike prior approaches relying on predefined gaits or demonstrations~\cite{drl:lee2020learning,hloc:bjelonic2020rolling,hloc:bjelonic2021whole,hmpc:kang2023rl+,hmpc:gangapurwala2022rloc,mpc:grandia2023perceptive,hmpc:jenelten2024dtc}, our method learns acyclic contact patterns directly through trial and error, while delegating motion execution to the MPC. This division of roles greatly simplifies reward design and enables the emergence of diverse, task-adaptive behaviors. The same MPC is used during training and deployment, ensuring robustness, sample efficiency and generalization across domains.
Our system demonstrates robust performance across platforms with different topologies and weight distributions ($50$--$120~\mathrm{kg}$), successfully transferring from simulation to real hardware, without domain randomization or fine-tuning. To support the proposed architecture, we develop a scalable software framework capable of running thousands of MPCs in parallel on CPU, while interfacing with GPU-based training. Policies are trained using Soft Actor Critic (SAC)~\cite{drl:haarnoja2018soft}, a design choice that further increases sample efficiency, without using demonstrations to aid the training. We validate the approach on flat-terrain tasks on legged and hybrid wheeled-legged locomotion, observing the emergence of non-periodic contact patterns and adaptive stepping behaviors. \revblock{We further demonstrate the extensibility of the architecture to unstructured environments by training a hybrid locomotion agent on a terrain composed of pyramidal steps.}

%% file: 02-related_works.tex
\section{Related work}\label{sec:rel_works}

\subsection{Model Predictive Control}
For complex robots and environments with multiple possible contacts, proper contact scheduling is critical~\cite{survey:wensing2023optimization}. In MPC-based locomotion, contact scheduling is typically addressed using one of three approaches.

\subsubsection{Predefined Contact Scheduling}
The optimal control problem is transcribed into a nonlinear program using a predefined contact sequence, with contact forces treated as decision variables and constrained by friction and unilaterality~\cite{hloc:bjelonic2020rolling,hloc:bjelonic2021whole,mpc:grandia2023perceptive}.

\subsubsection{Hybrid Optimization}
Hybrid formulations jointly optimize robot motion and discrete contact sequences, resulting in mixed-integer or bilevel optimization problems~\cite{survey:wensing2023optimization}. While expressive, these methods may require exploring a combinatorial number of contact sequences, often limiting real-time applicability.

\subsubsection{Contact-Implicit Formulations}
Contact-implicit methods enforce relations between system states and contact forces without discrete variables, typically via relaxed complementarity constraints~\cite{mpc:kim2024contact}. These approaches often rely on accurate initial guesses and are primarily suited for offline optimization~\cite{survey:wensing2023optimization}.

\subsection{Reinforcement Learning}
Unlike MPC, RL policies generally do not explicitly
model system dynamics. Instead, they map input observations directly to distributions over actions, learning contact-implicit policies that maximize future cumulative reward, implicitly handling nonsmooth contact dynamics. Compared with traditional model-based approaches which would typically require carefully engineered perception pipelines, end-to-end RL simplifies perception integration~\cite{drl:agarwal2023legged}. However, Deep-RL is typically sample inefficient and relies on massive parallel simulation for data collection~\cite{drl:rudin2022learning,rl:mjx_playground,drl:lee2020learning}. Domain randomization is commonly used to improve sim-to-real transfer, further increasing data requirements.
Model-based RL approaches can improve sample efficiency and enable real-world training via learned world models~\cite{drl:wu2023daydreamer}. However, the resulting policies are often difficult to interpret and transfer across domains, limiting their applicability to complex locomotion tasks.
\subsection{Hybrid Architectures}
Hybrid approaches combining learning and optimal control have been proposed to leverage the strengths of both paradigms. In~\cite{hmpc:kang2023rl+}, a simplified inverted pendulum model is used to generate reference motions that are imitated by a blind locomotion policy trained in simulation, with predefined contact schedules used at deployment. While effective for improving sample efficiency, such demonstration-based methods introduce biases tied to the simplified model.
Most hybrid methods instead adopt a hierarchical structure. In~\cite{hmpc:gangapurwala2022rloc}, a full rigid-body planner is coupled with an RL policy that predicts foothold locations from perception, using predefined contact schedules. Similarly,~\cite{hmpc:jenelten2024dtc} trains a tracking policy from planner rollouts for robust locomotion on sparse terrains, with the planner also active at deployment and enabling zero-shot transfer across planner instances. 
Alternative approaches address contact scheduling directly. In~\cite{hmpc:taouil2024non}, a Monte-Carlo Tree Search module replaces mixed-integer optimization and is combined with a simplified MPC and whole-body controller, with the MPC cost-to-go approximated via imitation learning for real-world deployment. Most closely related to our work is~\cite{hmpc:yang2022fast}, which learns a high-level gait policy using evolutionary strategies to parameterize contact phases for a convex MPC with simplified dynamics, showing that gait transitions can emerge from simple velocity tracking and energy-based rewards.

%% file: 03-overview.tex
\section{Overview}\label{sec:overview}
\begin{figure*}[t]
	\centering
	\includegraphics[width=0.95\textwidth]{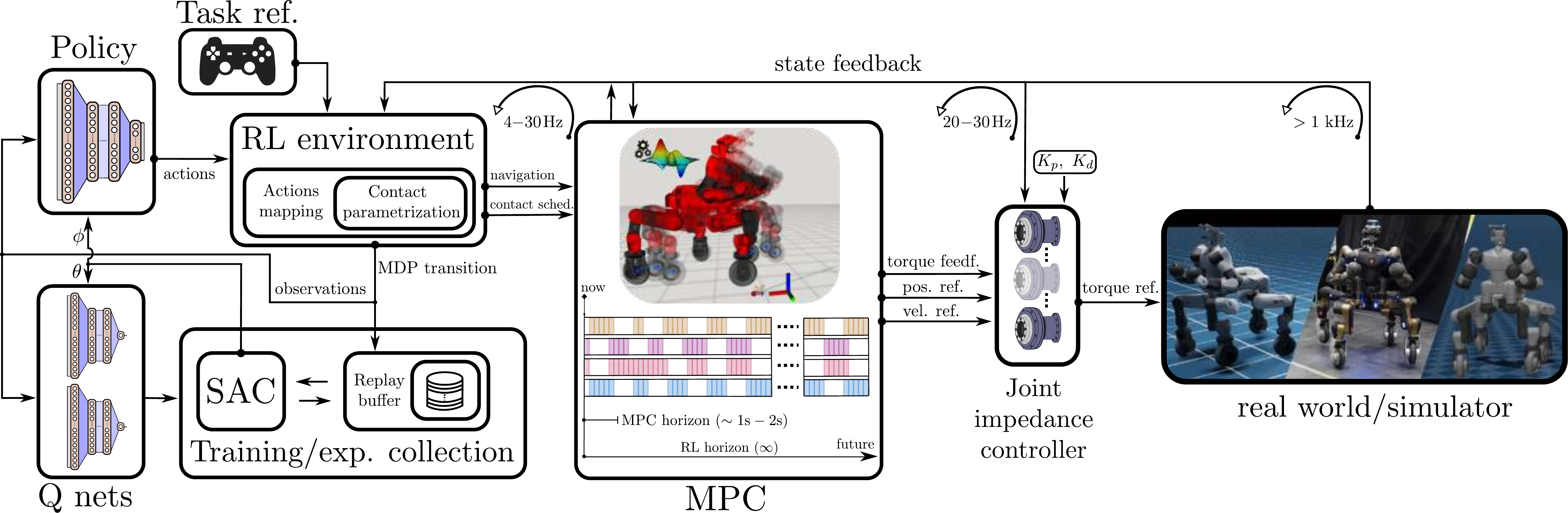}
	\caption{
	\revblock{The proposed hierarchical architecture uses an RL policy to generate both contact schedules and navigation commands for the underlying MPC.}
	}
	\label{fig:overview}
\end{figure*}

Our approach, illustrated in Fig.~\ref{fig:overview}, combines a low-level MPC locomotion controller with a high-level policy responsible for both contact scheduling and navigation. Unlike~\cite{hmpc:yang2022fast}, we train the policy using RL (SAC~\cite{drl:haarnoja2018soft}) and employ a full rigid-body dynamics model within the MPC. In contrast to approaches that leverage MPC-generated demonstrations~\cite{hmpc:kang2023rl+,hmpc:jenelten2024dtc}, the MPC and the scheduling policy operate concurrently and synchronously during both training and evaluation.
The MPC is formulated assuming a predefined contact schedule over the current optimization horizon~\cite{survey:wensing2023optimization}, enabling the use of efficient DDP-based solvers~\cite{to:mastalli2023inverse}. Our receding-horizon implementation allows flight phases to be injected on demand for each contact and their properties to be modified prior to each solver iteration (Sec.~\ref{subsec:horizon_control}). We further adopt an instantaneous parametrization of step actions, eliminating the need for clock-based observations while still allowing the policy to generate fully acyclic gaits through its interaction with the MPC (Sec.~\ref{sec:MDP}).
To ensure practical training efficiency, we developed a scalable and modular software framework capable of running thousands of MPC instances in parallel on CPU, while synchronizing with GPU-based simulation environments (Sec.~\ref{sec:soft_arch}).

%% file: 04-formulation.tex
\section{Low-level Model-based Control}\label{sec:MPC}
\subsection{Preliminaries}
We define the state and input spaces as $X\,\subseteq\,\mathbb{R}^{n_q + n_v}$ and $
U\,\subseteq\,\mathbb{R}^{n_u}$, where $n_q$ and $n_v$ denote the dimensions of the configuration and velocity spaces of the system, and $n_u$ the dimension of the input space. 
At each sample time $t_k$, given an optimization horizon of duration $N\cdot\Delta t$ 
, the low-level MPC controller solves a parametric NLP of the form:
\begin{align}\label{eq:nlocp}
\min_{\vc{x}_{0:N},\; \vc{u}_{0:N-1}} \quad & \sum_{i=0}^{N-1} \ell_i(\vc{x}_i, \vc{u}_i) + \ell_N(\vc{x}_N)\\
\text{s.t.} \quad &  \vc{x}_{i+1}=\vc{f}_i(\vc{x}_i, \vc{u}_i), \notag \\
& \vc{h}_i(\vc{x}_i, \vc{u}_i) = 0, \notag \\
& \vc{g}_i(\vc{x}_i, \vc{u}_i) \geq 0  \quad \forall i = 0, \dotsc, N-1 \notag \\
& \vc{g}_N(\vc{x}_N) \geq 0, \quad \vc{h}_N(\vc{x}_N) = 0 \notag
\end{align}
where $\ell_i\in \mathbb{R}$ and $\ell_N\in \mathbb{R}$ denote the running and terminal costs.
Dependency of~\eqref{eq:nlocp} upon the NLP parameters $\vc{\psi}\in\mathbb{R}^{n_p}$ was omitted for brevity. The constraints include the discrete dynamics $\vc{f}_i \in \mathbb{R}^{n_q+n_v}$, equality constraints $\vc{h}_i \in \mathbb{R}^{m}$, and inequality constraints $\vc{g}_i \in \mathbb{R}^{n}$.
The solution to~\eqref{eq:nlocp} yields a locally optimal state-input trajectory:
\begin{equation*}\label{eq:sol_set_tuple}
\mathcal{T}^{(k)}:= 
\bigl(\,
\vc{x}_0^{(k)},\dots,\vc{x}_N^{(k)},\;
\vc{u}_0^{(k)},\dots,\vc{u}_{N-1}^{(k)}
\,\bigr) 
\;\in\;
X^{N+1}\times U^{N}
\end{equation*}
We adopt an \textit{inverse-dynamics} formulation of~\eqref{eq:nlocp}, which has been shown to offer numerical advantages over forward-dynamics approaches~\cite{to:mastalli2023inverse} and enforce full rigid-body dynamics via a floating-base constraint $\left[\vc{\tau}_i\right]_{\mathrm{fb}}= \mathbf{0}\in\mathbb{R}^{6}$ for $i=0, ...,{N-1}$, where $\vc{\tau}_i\in\mathbb{R}^{n_v}$ are the generalized efforts.
The decision variables are defined as  $\vc{x}_i=\left(\vc{q}_i,\,\dot{\vc{q}}_i\right)$ and  $\vc{u}_i=\left(\ddot{\vc{q}}_i,\,\vc{\lambda}^{(i)}\right)$, where $\vc{\lambda}^{(i)}\in\mathbb{R}^{3\cdot n_c}$ is the concatenation of the $n_c$ foot contact forces at node $i$, and $\vc{q}_i\in\mathbb{R}^{n_q}$, $\dot{\vc{q}}_i\in\mathbb{R}^{n_v}$ and $\ddot{\vc{q}}_i\in\mathbb{R}^{n_v}$ denote the configuration, generalized velocity, and acceleration of the system. \revblock{The floating base orientation is encoded with a unit quaternion, while twists, accelerations, and contact wrenches are expressed in a world-aligned local body frame, which ensures that the floating-base dynamics are independent of joint torques, resulting in a compact and computationally efficient NLP.}

\subsection{Solution Approach and Stepping Logic}\label{subsec:mpc_step_logic}
Parametric NLPs such as~\eqref{eq:nlocp} can be solved efficiently using algorithms derived from DDP~\cite{to:mastalli2023inverse}. We employ an equality-constrained multiple-shooting variant of the Iterative Linear Quadratic Regulator (ILQR), available in~\cite{to:ruscelli2022horizon} \revblock{and approximate inequality constraints via quadratic barriers.}
At time $t_k$, the solver is initialized with the shifted previous solution $\mathcal{T}^{(k-1)}$ as initial guess $\mathcal{T}_{\mathrm{IG}}^{(k)}$ and $\vc{x}_{\mathrm{IS}}^{(k)}$ as initial state. A new solution $\mathcal{T}^{(k)}$ is then computed, following a real-time iteration scheme~\cite{mpc:diehl2005real}.
Due to the required computation time, the solution $\mathcal{T}^{(k)}$ becomes available only at the next time step $t_{k+1}$.
To compensate for this delay, prior to the solver call, we set the references for the underlying joint impedance controller based on the 
control input at the second node $\vc{u}_1^{(k-1)}$, instead of $\vc{u}_0^{(k-1)}$.
This adjustment anticipates the system's evolution during solver computation.
\subsection{Cost and Constraint Formulation}\label{subsec:formulation}
We focus on legged and hybrid wheeled-legged quadrupedal locomotion tasks. Our MPC framework employs an almost identical set of costs and constraints across all platforms, with only minor adaptations for hybrid locomotion. A detailed overview of the formulation is provided in Table~\ref{tab:mpc_form}, with the main cost weights reported in parentheses.
The problem includes equality constraints for numerical integration, $h_{\mathrm{int}}\in\mathbb{R}^{n_q+n_v}$, and inverse dynamics, $h_{\mathrm{dyn}}\in\mathbb{R}^{6}$. The initial state $\vc{x}_0$ is treated as an optimization variable, providing additional flexibility in closing the MPC state loop. For hybrid locomotion, rolling contact constraints $h_{\mathrm{rc}}^{(j)}\in\mathbb{R}^{3}$ are enforced on active contacts; otherwise, a no-slip point contact constraint $h_{\mathrm{pc}}^{(j)}\in\mathbb{R}^{3}$ is used.
The formulation further includes regularization costs on states and inputs, joint velocity limits $g_{\dot{q},\mathrm{lim}}\in\mathbb{R}^{n_{\mathrm{jnt}}}$, contact force unilaterality $g_{\mathrm{uni}}^{(j)}\in\mathbb{R}$, and friction cone constraints $g_{\mathrm{fr}}^{(j)}\in\mathbb{R}$ \revblock{(no joint torque limits are imposed)}. Terminal costs $l_{q,\mathrm{cap}}$ and $l_{\xi,\mathrm{cap}}$ promote desirable configurations at the end of the horizon.
Motion is guided by a base twist tracking cost $l_{\xi,\mathrm{ref}}(\ximpc)$ with $\ximpc\in\mathbb{R}^6$, together with vertical foot tracking costs $l_{v_z}^{(j)}(\feetvref)$ to encourage the tracking of desired foot trajectories. Contact schedules and timings are modified by activating or deactivating non-static costs and constraints (see Sec.~\ref{subsec:horizon_control}).

\begin{table}[t]
	\centering
	\caption{Problem costs/constraints formulation:\textit{ Grey-shaded costs are specific to the Centauro platform, whereas red and blue-shaded terms are used to define contact and flight phases, respectively.}}
	\resizebox{\columnwidth}{!}{
		\setlength{\tabcolsep}{3pt}
		\begin{tabular}{|c|c|l|c|c|c|}
			\hline
			 \textbf{Name} & \textbf{Descr.} & \textbf{Nominal Expr.} & \textbf{Dim.} & \textbf{Span} & \textbf{Support} \\
			\hline \hline 
			$h_{\mathrm{int}}$ & integrator  & $\vc{f}_i\,\ominus\,\vc{x}_{i+1} = 0$ & $n_q{+}n_v$ &  $N$ & $0{:}N{-}1$ \\ \hline
			$h_{\mathrm{dyn}}$ & inv. dynamics  & $[\vc{\tau}_i]_{\mathrm{fb}}=0$ & 6 &   $N$ & $0{:}N{-}1$ \\
			\hline
			$h_{x_\mathrm{init}}$ & init. state  & $[\vc{x}_{0}\,-\,\vc{x}_{\mathrm{IS}}]_{a:b} = 0$ & $n_{\mathrm{init}}$ &   $1$ & $0$ \\
			\hline
			\rowcolor{cyan!25}
			$h_{\mathrm{liftoff}}^{(j)}$ & force vanishing & $ \vc{\lambda}^{(i)}_j= 0$       & $3$ &  variable &  variable\\
			\hline
			\rowcolor{magenta!10}
			$h_{\mathrm{pc}}^{(j)}$ & no-slip contact & $ \vc{v}_{c, i}^{(j)}= 0$  & $3$ &  variable &  variable\\
			\hline
			\rowcolor{magenta!10}
			$h_{\mathrm{rc}}^{(j)}$ & no-slip rolling contact  & $\vc{v}^{(j)}_{w, i}{+}  R_i{\cdot}\vc{\omega}_i^{(j)}{\times} \hat{\vc{r}}_i^{(j)}{=}0$                 & $3$ & variable &  variable\\
			\hline\hline
			\rowcolor{magenta!10}
			$g_{\mathrm{fr}}^{(j)}$    & friction cone & $ \|\vc{\lambda}_{j, t}^{(i)}\| \leq \mu\cdot \lambda_{j, n}^{(i)} $    & $1$ &   variable &  variable\\
			\hline
			\rowcolor{magenta!10}
			$g_{\mathrm{uni}}^{(j)}$    & unilaterality & $ \lambda_{j, n}^{(i)} \geq 0$  & $1$ & variable & variable\\
			\hline
			$g_{\dot{q},\mathrm{lim}}$    & joint vel. limits  & $ [\dot{\vc{q}}_{\mathrm{lb}}]_{6:}{\leq} [\dot{\vc{q}}_i]_{6:}{\leq} [\dot{\vc{q}}_{\mathrm{ub}}]_{6:}$  & $n_{\mathrm{jnt}}$ & $N$ & $0{:}N{-}1$\\
			\hline
			\hline
			$l_{\dot{q}, \mathrm{reg}}$    & vel. regularization  & $ \|\dot{\vc{q}}_i\|^2$  & 1 & $N$ &  $0{:}N{-}1$\\
			\hline
			$l_{a, \mathrm{reg}}$    & acc. regularization  & $ \|\ddot{\vc{q}}_i\|^2$  & 1 & $N$ &  $0{:}N{-}1$\\
			\hline
			\rowcolor{magenta!10}
			$l^{(j)}_{\lambda}(\hat{\vc{\lambda}}_{j})$    & force regularization  & $ \|\vc{\lambda}_j^{(i)} - \hat{\vc{\lambda}}^{{(i)}}_{j}\|^2$  & $1$ & variable &  variable\\
			\hline
			$l_{\xi, \mathrm{cap}}$    & base capture  & $ \|\,[\dot{\vc{q}}_i]_{0:6}\,\|^2$  & 1 & $N/6$ &  $5/6N{:}N{-}1$\\
			\hline
			$l_{q, \mathrm{cap}}([\hat{\vc{q}}]_{7:})$    & posture capture  & $ \|\,[\vc{q}_i]_{7:}-[\hat{\vc{q}}]_{7:}\,\|^2$  & 1 & $N/6$ &  $5/6N{:}N{-}1$\\
			\hline
			$l_{\xi, \mathrm{ref}}(\ximpc)$    & base twist tracking  & $ \|\,[\dot{\vc{q}}_i]_{0:6}-\ximpc\,\|^2$  & 1 & $5/6N$ &  $0{:}5/6N{-}1$\\
			\hline
			\rowcolor{cyan!25}
			$l^{(j)}_{v_z}(\feetvref)$    & foot tracking  & $ \|[\vc{v}^{(j)}_{\mathrm{ee},i}]_{2} - {\vc{v}}^{{(j,\,i)}}_{\mathrm{ref},\,z}\|^2$  & $1$ & variable &  variable\\
			\hline
			\rowcolor{gray!15}
			\cellcolor{cyan!25}
			$l_{\mathrm{vert}}^{(j)}$   & vertical touchdown  & $ \|[\vc{v}_{\mathrm{ee},i}^{(j)}]_{0:2} \|^2$  & $1$ & 1 &  variable\\
			\hline
			\rowcolor{gray!15}
			$l_{\mathrm{ee,ori}}^{(j)}$   & foot orientation  & $ \left\| \vc{R}_{\mathrm{ee},i}^{(j)} \hat{\vc{z}} - \hat{\vc{z}} \right\|^2 $  & $1$ & $N$ &  $0{:}N{-}1$\\
			\hline
	\end{tabular}}
	\noindent\parbox{\columnwidth}{
		\scriptsize{
			\vspace{0.2cm}
			\revblock{For $l^{(j)}_{\lambda}(\hat{\vc{\lambda}}_j)$ we target equal load sharing: for each active contact, the normal force is regularized around $m\!\cdot\!g/n_{\mathrm{ac}}$ ($n_{\mathrm{ac}}$ number of active contacts; tangential target is $0$). $\hat{\vc{q}}$ in $l_{q,\mathrm{cap}}$ corresponds to the static equal-load posture when all contacts are active.}
		}
	}
	\label{tab:mpc_form}
\end{table}

\subsection{Horizon Control and Acyclic Contact Scheduling}\label{subsec:horizon_control}
\revblock{Formally, we group cost/constraint terms over the horizon into distinct \textit{phases} $\mathcal{P}^{(j)}$ (per foot $j$). As shown in Table~\ref{tab:mpc_form}, some terms have non-static support (i.e., the subset of horizon nodes on which they are active) and node span (phase duration in nodes), which can be modified at runtime.} 
We define two phase types: \textit{flight phases} $\mathcal{P}_{\mathrm{fl}}^{(j)}$, where $h_{\mathrm{liftoff}}^{(j)}$ and $l_{\mathrm{ee}, v_z}^{(j)}$ are active, and \textit{contact phases} $\mathcal{P}_{\mathrm{c}}^{(j)}$, where $h_{\mathrm{pc}}^{(j)}$, $h_{\mathrm{rc}}^{(j)}$, $g_{\mathrm{fr}}^{(j)}$, $g_{\mathrm{uni}}^{(j)}$, and $l_{\lambda}^{(j)}$ are active (see Table~\ref{tab:mpc_form}). 
\revblock{
	At each instant $t_k$, for each foot independently, a scalar injection action $\chimpcj \in \mathbb{R}$ controls whether a new flight phase $\mathcal{P}_{\mathrm{fl}}^{(j)}$ of length $T_{\mathrm{fl}}^{(j)} \in \mathbb{Z}_{>0}$ should be injected; in practice, an injection is triggered when $\chimpcj < 0$.
}
Injections are only allowed at a fixed \revblock{injection node $i^*$},
and only if there is no overlap with a pre-existing flight phase for the same foot. Since the MPC operates under a real-time iteration \revblock{(RTI)} regime~\cite{mpc:diehl2005real}, the choice of $i^{*}$ reflects a trade-off between responsiveness and the feasibility of the NLP. \revblock{Choosing a smaller $i^*$ reduces injection latency ($\approx i^*\Delta t$) but leaves fewer horizon nodes to accommodate the new flight phase and satisfy constraints under RTI, increasing the risk of infeasibility/poor convergence; larger $i^*$ improves feasibility at the cost of delayed reaction.} To maintain a fully populated horizon when no injections occur, unit-length $\mathcal{P}_{\mathrm{c}}^{(j)}$ are continuously appended at the end of the horizon. 

With this formulation, the current contact schedule 
for the $j$\text{-th} foot is fully described by the ordered sequence of the $n_{\mathcal{P}_{\mathrm{fl}}}^{(j)}$ flight phases $\mathcal{P}_{\mathrm{fl}}^{(j,\,m)}$ present over the horizon, with \revblock{$m = 0, \dots, n^{(j)}_{\mathcal{P}_{\mathrm{fl}}} - 1$}. Each $\mathcal{P}_{\mathrm{fl}}^{(j,\,m)}$ is characterized by its current position \revblock{$d_{\mathrm{fl}}^{(j,\,m)} \in \mathbb{Z}_{\geq 0}$}, current node span \revblock{$s_{\mathrm{fl}}^{(j,\,m)} \in \mathbb{Z}_{> 0}$}, and vertical foot reference trajectory $\feetvrefk \in \mathbb{R}^{s_{\mathrm{fl}}^{(j,\,m)}}$. Instead of specifying $\feetvrefk$ on each node, we compute it from a cubic polynomial trajectory parametrized by a desired clearance $H_{c}^{(j,\,m)}\in \mathbb{R}$ and a landing height difference $H_{l}^{(j, m)}\in \mathbb{R}$.
\subsection{Closing the State Loop on the MPC}\label{subsec:closing_loop}
\revblock{The MPC closed-loop behavior is set by the initial-state constraint $\vc{x}_0=\vc{x}_{\mathrm{IS}}^{(k)}$: open-loop (OL) uses $\vc{x}_\mathrm{IS}^{(k)}=\vc{x}_1^{(k-1)}$ (previous solution), full closed-loop uses $\vc{x}_\mathrm{IS}^{(k)}={\vc{x}}^{k}_{\mathrm{meas}}$ (fully observed state), and partial closed-loop (CLP) constrains only measured components of $\vc{x}_0$. We use OL or CLP (IMU and joint encoders only); in OL we replace $l_{v_z}^{(j)}$ with a position-based foot tracking cost $l_{p_z}^{(j)}$ to avoid contact drift.}
\section{High-level Control: navigation and contact scheduling}\label{sec:MDP}
Our MPC requires as input a contact schedule and base twist commands. To generate and adapt these inputs according to task demands, we leverage a high-level RL policy trained in simulation.
\subsection{Task Formulation} \label{subsec:rl_task_descr}
\revblock{In our main evaluation, we focus on an infinite-horizon base twist tracking task, using flat terrain as a baseline setting in which proper contact scheduling remains non-trivial, particularly during direction changes and transitions between wheeled and legged locomotion. This setting allows us to clearly demonstrate the proposed architecture’s ability to discover non-periodic contact patterns, while the approach naturally extends to more complex scenarios, as shown in Sec.~\ref{subsec:nonflat}}.

In our problem formulation, the task goal is specified as a target position in the plane $\vc{p}^{\mathrm{cmd}} \in \mathbb{R}^2$, rather than a target base twist, as is more commonly done in other works~\cite{rl:mjx_playground}. Based on the error w.r.t. $\vc{p}^{\mathrm{cmd}}$, at each instant, we compute a corresponding desired twist $\vc{\xi}^{\mathrm{cmd}} \in \mathbb{R}^6$ according to a linear ramp. $\vc{\xi}^{\mathrm{cmd}}$ is both observed by the policy and used for reward computation. This indirect formulation retains the simplicity of twist tracking, while resulting in more informative and realistic experience, aligning better with the demands of real-world deployment. We found that this approach, especially for hybrid locomotion, encourages more effective stepping behaviors. 
\subsection{Markov Decision Process Formulation}\label{subsec:mdp}
The task described in Sec.~\ref{subsec:rl_task_descr} can be naturally formulated as a Markov Decision Process (MDP). 
We define the policy to be hierarchically coupled with the MPC controller. We define the MDP actions $\vc{a}\in\mathcal{A}$ to be the MPC inputs $\ximpc$ and $\chimpcj$ (see Secs.~\ref{subsec:formulation} and~\ref{subsec:horizon_control}) corresponding to navigation and contact schedule control.
The MPC loop runs at a frequency $f_{\mathrm{MPC}}{=}1/\Delta t$, while the RL policy runs at slower frequency $f_{\pi}{=}\,f_{\mathrm{MPC}}/n_{\mathrm{rep}}$, where $n_{\mathrm{rep}}\in \mathcal{N}^{+}$ is the number of MPC steps per policy action.
The resulting state $\vc{s} \in \mathcal{S} = \mathcal{S}_{\mathrm{task}} {\times} \mathcal{S}_{\mathrm{robot}} {\times} \mathcal{S}_{\mathrm{MPC}}$ is composed of the goal state, robot state, and the full MPC state, respectively. Both $\mathcal{S}_{\mathrm{task}}$ and $\mathcal{S}_{\mathrm{MPC}}$ are, in principle, fully observable. However, using the full MPC state as an observation is impractical in real-world settings, and the robot state $\mathcal{S}_{\mathrm{robot}}$ is typically not fully known, making the environment partially observable in practice. As a result, we define the observation $\vc{o}\in\mathcal{O}$, with $\mathcal{O}{=}\mathcal{O}_{\mathrm{task}}{\times}\mathcal{O}_{\mathrm{robot}}{\times} \mathcal{O}_{\mathrm{MPC}}{\times}\mathcal{O}_{\mathrm{IMP}}{\times}\mathcal{O}_{\mathrm{act}}$, integrating information from the task, robot proprioception, MPC, joint impedance controller and action history. With this formulation we define the policy as a stochastic distribution $\pi(\vc{a} \mid \vc{s})$ and adopt an entropy-regularized RL objective~\cite{rl:ziebart2010modeling} to encourage exploration and multimodal behavior.
\subsubsection{Action Space}\label{subsubsec:action_space}
The policy controls the MPC twist reference $\ximpc\in[\xi_{\mathrm{lb}}, \xi_{\mathrm{ub}}]^6$ and the current contact schedule by injecting new flight phases, through the injection actions \revblock{$\chimpc\in \mathbb{R}^{n_c}$} (see Sec.~\ref{subsec:horizon_control}), where $n_c$ is the number of feet. The resulting action space is then $\mathcal{A}\subseteq\mathbb{R}^{6+n_c}$, with $\vc{a}\triangleq\left(\ximpc,\chimpc\right)$.
\subsubsection{Observation Space}
The proprioceptive data $\mathcal{O}_{\mathrm{robot}}$ includes base-local observations of the normalized gravity vector $\hat{\vc{g}}_n\in\mathbb{R}^{3}$, angular velocity $\vc{\omega}\in\mathbb{R}^{3}$, joint positions and velocities. The current twist command $\vc{\xi}^{\mathrm{cmd}}\in\mathbb{R}^{6}$ is also observed in a base-local frame. We assume no odometry or contact sensing is available on the robot and estimate linear velocity $\vc{v}\in\mathbb{R}^{3}$ and contact forces $\vc{\lambda}\in\mathbb{R}^{3\cdot n_c}$ based on the last MPC solution $\mathcal{T}^{(k-1)}$. Instead of observing the full MPC state--which is impractical--we construct an observation subset, shown in Table~\ref{tab:obs_def}, that captures the minimal MPC information we deem sufficient for task completion. This includes the duration and position of the earliest flight phase for each contact, which vary as phases shift over time.
Additionally, we include \revblock{a merit-function-inspired} \enquote{health} index $\delta_{\mathrm{MPC}}\in\mathbb{R}$, computed as a smoothed version of  $l(\mathcal{T}^{(k-1)}){+}\kappa \cdot \Vert \vc{h}(\mathcal{T}^{(k-1)}) \Vert_1$, where $l$ denotes the total cost, $\vc{h}\in\mathbb{R}^{m}$ is the constraints residual, and $\kappa$ is a scaling factor \revblock{($\delta_{\mathrm{MPC}}\approx0$: well converged/feasible; $\delta_{\mathrm{MPC}}\gg0$: solver divergence/constraint violation)}.
The observation also includes the reference joint position, velocity and feedforward torques for the joint impedance controller. In addition, a history of past actions $\mathcal{O}_{\mathrm{act}}{\subseteq} \mathbb{R}^{(6+n_c)\cdot n_h}$ of length $n_h$ is provided.
Notably, due to the adopted action parametrization, no clock-based information is required as part of the observation. 
\begin{table}[t]
	\centering
	\renewcommand{\arraystretch}{1.3}
	\caption{Observations from MPC: \textit{The agent observes linear velocity and contact forces estimates from the MPC, a minimal representation of its convergence status and the current contact schedule.}}
	\resizebox{\columnwidth}{!}{
		\setlength{\tabcolsep}{3pt}
		\begin{tabular}{|c|c|c|l|c|c|c|}
			\hline
			\textbf{Name} & \textbf{Descr.} & \textbf{Type} & \textbf{Range} & \textbf{Dim.} & \textbf{Obs. subspace} & \textbf{Ref. frame}\\
			\hline \hline
			$\vc{v}$ & linear velocity (prediction) & real &$[-\infty, \infty]$ & $3$ & $\mathcal{O}_{\mathrm{MPC}}$ & base loc.\\ \hline
			$\vc{\lambda}$ & contact forces (prediction) & real &$[-\infty, \infty]$ & $3\cdot n_c$ & $\mathcal{O}_{\mathrm{MPC}}$ & base loc.\\ \hline
			$\delta_{\mathrm{MPC}}$ & MPC health $\propto (l +\kappa\cdot\Vert \vc{h}\Vert_1)$ & real &$[0, \infty]$ & $1$ & $\mathcal{O}_{\mathrm{MPC}}$ & $\cdot$\\ \hline
			$d_{\mathrm{fl}}^{(:, 0)}$ & $\mathcal{P}_{\mathrm{fl}}^{(:, 0)}$ \textit{current} position & integer &$[0, i^{*}]$ & $n_c$ & $\mathcal{O}_{\mathrm{MPC}}$ & $\cdot$\\ \hline
			$s_{\mathrm{fl}}^{(:, 0)}$ & $\mathcal{P}_{\mathrm{fl}}^{(:, 0)}$ \textit{current} node span& integer & $[0, N]$ & $n_c$ & $\mathcal{O}_{\mathrm{MPC}}$ & $\cdot$\\ \hline
		\end{tabular}
	}
	\label{tab:obs_def}
\end{table}

\subsubsection{Rewards}\label{subsubsec:rewards}
The reward $r(t_k)$ is defined as a weighted sum of three components. 

\revblock{The tracking term encourages the policy to follow the high-level command $\vc{\xi}^{\mathrm{cmd}}$.
	Letting $\bar{\vc{\xi}}{=}\left(\bar{\vc{v}},\,\bar{\vc{\omega}}\right)$ and $\vc{\xi}^{\mathrm{cmd}}{=}\left(\vc{v}^{\mathrm{cmd}},\,\vc{\omega}^{\mathrm{cmd}}\right)$ denote, respectively, the average and commanded twist over the step, we define $\Delta{\vc{v}}{\triangleq}\bar{\vc{v}}{-}\vc{v}^{\mathrm{cmd}}$, $\Delta{\vc{\omega}}{\triangleq}\bar{\vc{\omega}}{-}\vc{\omega}^{\mathrm{cmd}}$, and $\Delta\vc{\xi}{\triangleq}\left(\tilde{v}_x,\,\tilde{v}_y,\,\tilde{v}_z,\,\Delta{\vc{\omega}}\right)$, where the projected velocity error components are defined as $\tilde{v}_z{=}\Delta{\vc{v}}{\cdot}\hat{\vc{g}}_n$, $\tilde{v}_x{=}\Delta{\vc{v}}{\cdot}\hat{\vc{v}}^{\mathrm{cmd}}$, and
	$\tilde{v}_y{=}\bigl\Vert\Delta{\vc{v}}{-}\tilde{v}_z\hat{\vc{g}}_n{-}\tilde{v}_x\hat{\vc{v}}^{\mathrm{cmd}}\bigr\Vert$, with $\hat{\vc{v}}^{\mathrm{cmd}}{=}\frac{\vc{v}^{\mathrm{cmd}}}{\Vert\vc{v}^{\mathrm{cmd}}\Vert}$.}
Following these definitions, we construct the tracking reward as
\begin{align*}
	&r_{\xi}=w_{\xi}\cdot e^{-\kappa_{\xi}{
			\tfrac{\Vert \Delta\,{\vc{\xi}}\circ\vc{w}^{'}_\xi\Vert}{\|\vc{w}^{'}_\xi\|}
	}},\;\vc{w}^{'}_\xi{=}\left(1,\sqrt{0.1},\sqrt{0.1},0,0,1.0 \right).
\end{align*}

This formulation allows emphasizing tracking along specific axes, which we found particularly important for slow-reacting platforms. Unless stated otherwise, we use $w_{\xi}=1.0$ and $\kappa_{\xi}=5.0$.
The second reward term penalizes the rate of change of the agent's actions:
\revblock{\begin{equation*}
r_{a}=w_{a}\!\left(1-\kappa_{a}\!\left(\|\tilde{\vc{a}}_{\xi}\|^{2}+0.1\|\tilde{\vc{a}}_{\chi}\|^{2}\right)\right)
\end{equation*}
where $\tilde{\vc{a}}_{\xi} $ and $\tilde{\vc{a}}_{\chi}$ are the action components referring to the twist $\ximpc$ and the contact flags $\chimpc$, respectively.} We use $w_{a}{=}0.1$ and $\kappa_{a}{=}2.0$.
The third reward term 
\begin{align*}\label{eq:cotrew}
	&r_{\mathrm{CoT}}=w_{\mathrm{CoT}}\cdot\left[1-\kappa_{{CoT}}\cdot \frac{\drainedpow}{m\,g\,\Vert \vc{\xi}^{\mathrm{cmd}} \Vert +\epsilon}\right]
\end{align*}
is based on the average Cost of Transport (CoT) over each environment step and promotes energy-efficient behaviors. For this term, we use $w_{\mathrm{CoT}} = 0.1$ and $\kappa_{\mathrm{CoT}} = 1.0$. The terms $m$, $g$ and $\epsilon$ are the robot’s total mass, gravitational acceleration magnitude and a small constant, respectively.
\subsubsection{Episodes Definition}\label{sec:epdef}
To promote experience diversity and improve computational efficiency, training episodes are truncated at a maximum duration of $256/n_{\mathrm{rep}}$ steps. At each truncation, the target planar position $\vc{p}^{\mathrm{cmd}}$ is uniformly resampled within a $5\,\mathrm{m}$ radius around the robot. $\vc{\xi}^{\mathrm{cmd}}$ is continuously updated using a linear ramp based on the error w.r.t. $\vc{p}^{\mathrm{cmd}}$, and its norm is clipped to a maximum value ${v}^{\mathrm{max}}$.
Episodes terminate early if: (i) the robot capsizes, or (ii) the MPC \enquote{health} index $\delta_{\mathrm{MPC}}$ exceeds a threshold, indicating divergence. Truncations do not reset the simulation state, aiding exploration and encouraging robustness to varying conditions, while terminations always trigger a full reset. Resets place the robot in a safe pose with a random base yaw and reinitialize the MPC to its default state. 
\subsection{\revblock{Training Algorithm}}
In our setup, transition sampling from the underlying MDP is particularly expensive: the MPC operates at a low frequency compared to the physics simulation, resulting in a low number of environment steps-per-second. For this reason, we adopt an off-policy RL approach and employ the SAC algorithm~\cite{drl:haarnoja2018soft}, in combination with parallel experience collection.
\section{\revblock{Software Architecture}}\label{sec:soft_arch}
\begin{figure}[ht]
	\centering
	\includegraphics[width=0.98\columnwidth]{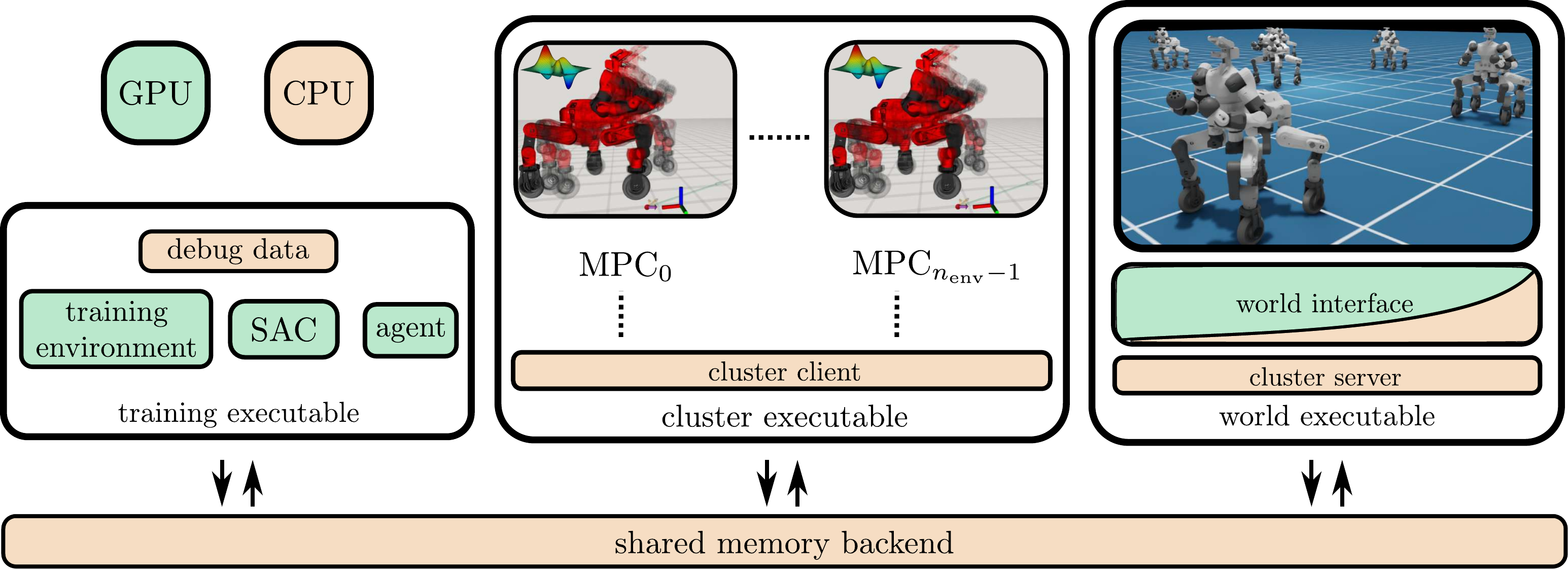}
	\caption{\revblock{High-level software architecture. The system is modular, with three main components: \textit{world} interface, MPC \textit{cluster} and \textit{training} environment. Green modules operate on GPU-resident data; orange modules run on CPU.}}
	\label{fig:soft_arch}
\end{figure}

Recent advances in robot learning have highlighted the importance of massively parallel experience collection to train robust and effective policies~\cite{drl:rudin2022learning}. To this end, we developed a modular and scalable software architecture that supports thousands of CPU-based MPC instances while integrating seamlessly with GPU-accelerated simulation. Figure~\ref{fig:soft_arch} provides a high-level overview of the architecture, which comprises three core components: the \textit{world} simulation, the MPC \textit{cluster}, and agent \textit{training}. This modular design improves maintainability and debugging flexibility, while introducing limited overhead.

While GPU-based vectorized simulators are now widely available, mature libraries for vectorized MPC remain limited~\cite{jeon2024cusadi}. To address this gap, we implement a custom CPU-based MPC parallelization library built on the Horizon framework~\cite{to:ruscelli2022horizon}, where each MPC instance runs in a separate subprocess coordinated via shared memory. The \textit{world} module provides a vectorized interface to the robot and the MPC cluster, supporting training in IsaacSim~\cite{sim:nvidia_isaacsim}, evaluation in MuJoCo~\cite{sim:deepmind2021mujoco}, and real-world deployment via the XBot2 middleware~\cite{laurenzi2023xbot2}. The \textit{training} module manages the RL loop, including the SAC agent and environment definition. MPC solutions are computed in parallel with physics integration, enabling the modeling of real-world action delays (see Sec.~\ref{subsec:mpc_step_logic}).

%% file: 05-results.tex
\section{Results}\label{sec:results}
We validate the proposed architecture and software framework through a series of experiments on multiple robot platforms with varying morphologies and weight distributions, covering both legged and wheeled-legged locomotion tasks on flat terrain. All policies are trained in simulation without domain randomization, using the same reward formulation (Sec.~\ref{subsubsec:rewards}) across experiments. We evaluate zero-shot sim-to-sim transfer on all platforms and demonstrate zero-shot sim-to-real transfer on Centauro. Experimental evaluations are conducted on a simplified $50\,\mathrm{kg}$ quadruped in both legged and wheeled configurations, a Unitree B2-W wheeled quadruped ($\approx 80\,\mathrm{kg}$), and Centauro, our $120\,\mathrm{kg}$ wheeled-legged humanoid robot (see Fig.~\ref{fig:intro}). \revblock{We also demonstrate an extension of the architecture to hybrid locomotion on non-flat terrain.} Videos of all experiments are available at \url{\videoURL}.

\subsection{Controller Hyperparameters}
Each robot runs a tailored instance of the MPC described in Sec.~\ref{sec:MPC}, with only minor platform-specific modifications. For Centauro, we deploy two MPC configurations: one for standard legged locomotion without wheel actuation, and one for hybrid wheeled-legged locomotion, both without upper-body control. In the hybrid configuration, control of the ankle omnidirectional steering joints is disabled to encourage the emergence of step reflexes.

Unless stated otherwise, all MPCs operate in partial closed-loop mode (see Sec.~\ref{subsec:closing_loop}) with a prediction horizon of $N=30$. The MPC time step is set to $\Delta t = 0.03\,\mathrm{s}$ and \revblock{$i^*=4$} (see Sec.~\ref{subsec:horizon_control}) for all platforms, except for Centauro, where a coarser $\Delta t = 0.05\,\mathrm{s}$ is used.
\revblock{For flat-terrain experiments, we fix $H_{c}^{(j,\,\cdot)} = 0.1\,\mathrm{m}$, $H_{l}^{(j,\,\cdot)} = 0\,\mathrm{m}$, and set $T_{\mathrm{fl}}^{(j,\,\cdot)}$ to $0.8\,\mathrm{s}$ for Centauro and to $0.6\,\mathrm{s}$ for all other platforms.}

\subsection{\revblock{Training and Evaluation Hyperparameters}}\label{subsec:hyper_params_rl}
The action space has fixed dimension $n_{\mathrm{a}} = 10$ across all study cases. The observation dimension scales with the number of joints, reaching up to $n_{\mathrm{obs}} = 250$ for Centauro. All trainings employ $n_{\mathrm{env}} = 800$ parallel environments and rely on extensive experience reuse. Each policy update uses batches of $16384$ transitions sampled from a replay buffer of size $n_{\mathrm{buffer}} = 15 \times 256 / n_{\mathrm{rep}} \times n_{\mathrm{env}}$.
Actor and critic networks consist of three hidden layers with constant width, set to $60\%$ of the input dimension. Finally, the maximum task velocity norm $v^{\mathrm{max}}$ (see Sec.~\ref{sec:epdef}) is set to $1\,\mathrm{m/s}$ for all platforms except Centauro, for which it is limited to $0.5\,\mathrm{m/s}$.
\subsection{\revblock{Framework Performance}}
\begin{figure}[ht]
	\centering
	\vspace{0.3cm}
	\includegraphics[width=0.98\columnwidth]{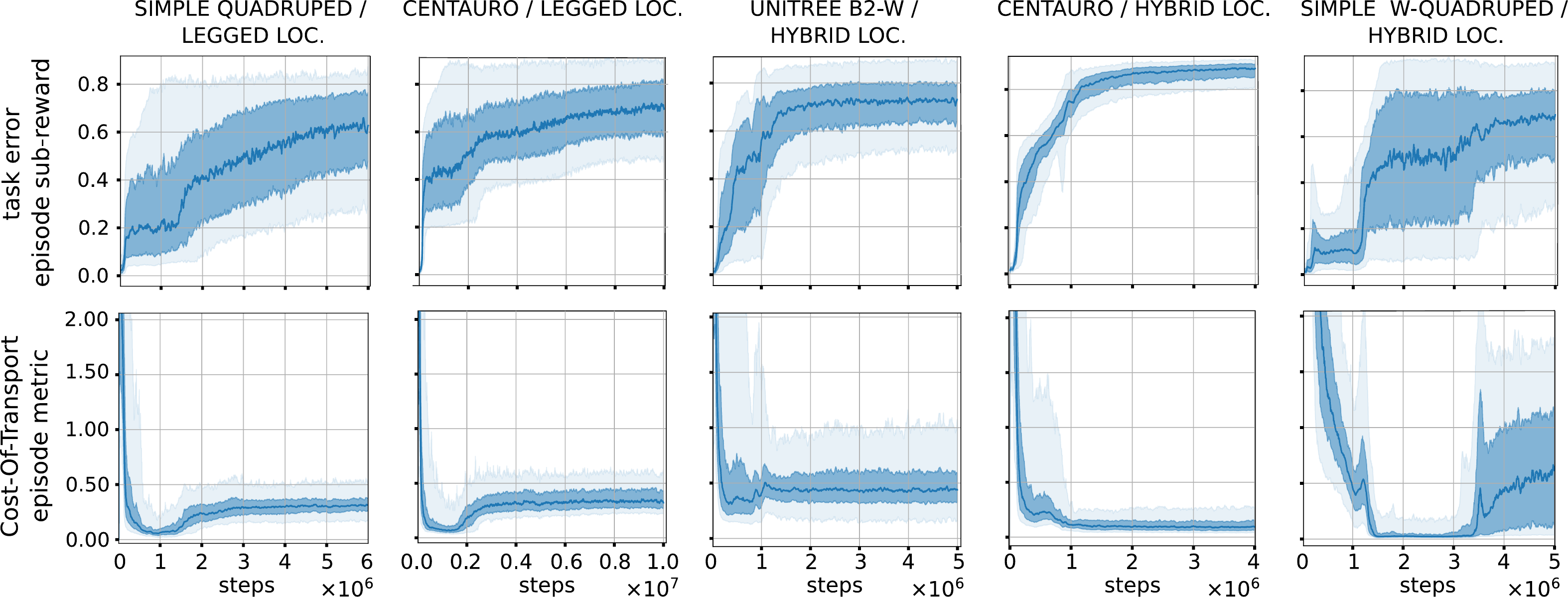}
	\caption{Task tracking episode sub-reward (first row) and CoT metric (second row) during training for all the flat terrain case studies, averaged across 800 environments. Dark shaded areas indicate the first and third quartiles of the distribution across environments; lighter regions denote the $5\text{-th}$ and $95\text{-th}$ percentiles. Action rate sub-rewards are omitted for brevity.} 
	\label{fig:sub_rewards}
\end{figure}

Figure~\ref{fig:sub_rewards} summarizes the training performance across all tested platforms for both legged and hybrid locomotion tasks on flat terrain. Across scenarios, policies are obtained within $4$–$10\times10^6$ environment steps, corresponding to approximately $9$–$29$ simulated days. By comparison, blind end-to-end RL policies typically require similar simulated time but consume an order of magnitude more samples~\cite{rl:mjx_playground}, highlighting the improved sample efficiency of the proposed approach.
During training, the framework achieves real-time factors of up to $50$, even on complex platforms such as Centauro \revblock{($n_v=43$)}, comparable to other hybrid approaches~\cite{hmpc:jenelten2024dtc}. Experiments were conducted on a workstation equipped with an AMD Ryzen Threadripper 7970 CPU, 128\,GiB RAM, and an NVIDIA RTX 4090 GPU, where $12$ simulated days of training correspond to approximately $5.8$ hours of wall time. While this wall time efficiency does not match that of fully end-to-end RL methods, the proposed approach offers improved robustness and real-world transferability (Sec.~\ref{subsec:transfer}). Once trained, policy inference incurs negligible overhead: for Centauro, RL evaluation requires only $0.334\,\mathrm{ms}$ on the onboard Intel Core i7-11800H CPU, \revblock{while the MPC runs in real time.}
\subsection{\revblock{Legged and Hybrid Wheeled-Legged Locomotion}}
\begin{figure*}[t]
	\centering
	\includegraphics[width=0.98\textwidth]{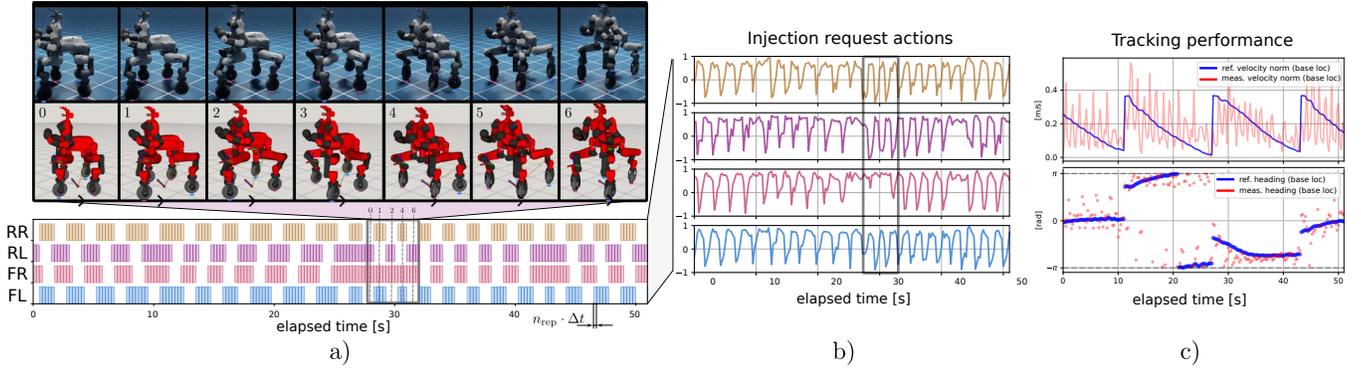}
	\caption{Deterministic evaluation of a legged locomotion policy for Centauro, shown over a $50\,\mathrm{s}$ window. The injection requests actions $\chimpc$  chosen by the policy (b) generate the contact schedule and motion (a), revealing completely acyclic contact patterns and timing adaptations. The associated tracking performance is shown in (c). Flight phases are injected within the MPC horizon, for each foot $j$, only when $\chimpcj<0$.} 
	\label{fig:acyclic_contacts_cloop}
\end{figure*}

To showcase the architecture’s ability to produce adaptive, acyclic gait patterns, we report a deterministic rollout of the trained legged locomotion policy on Centauro (Fig.~\ref{fig:sub_rewards}, second column). The evaluation employs cyclic triangular position waypoints that induce frequent changes in direction. \revblock{Importantly, although the terrain is flat, the task is inherently non-stationary due to time-varying references in both direction and magnitude (see Fig.~\ref{fig:acyclic_contacts_cloop}(c)), for which strictly periodic gaits are not expected to be optimal.}

The policy is trained and evaluated at $4\,\mathrm{Hz}$ for the SAC agent ($n_{\mathrm{rep}}{=}5$) and $20\,\mathrm{Hz}$ for the MPC. Figure~\ref{fig:acyclic_contacts_cloop} shows the resulting behavior over a $50\,\mathrm{s}$ window, where reference changes trigger distinctly acyclic contact sequences with alternating single and double flight phases. While a trotting pattern is typically employed during steady forward motion, the contact schedule adapts continuously: as the norm of the commanded twist $\vc{\xi}^{\mathrm{cmd}}$ decreases near the target position $\vc{p}^{\mathrm{cmd}}$, contact durations increase, corresponding to a smooth reduction in gait frequency. In addition to symmetric trotting, asymmetric trotting patterns also emerge.

The architecture was also trained on hybrid wheeled-legged locomotion across three platforms. In all cases, the policies primarily rely on the wheels to track the commanded velocity $\vc{\xi}^{\mathrm{cmd}}$, while stepping behaviors emerge when needed to reorient the robot before resuming wheeled motion (see supplementary material). Despite being trained with a ${v}^{\mathrm{max}}$ twice that of the legged policies (Sec.~\ref{subsec:hyper_params_rl}), the hybrid policies achieve smoother tracking and faster convergence, with lower variance (Fig.~\ref{fig:sub_rewards}).
\subsection{Policy Frequency Influence}
\begin{figure}[ht]
	\centering
	\includegraphics[width=0.98\columnwidth]{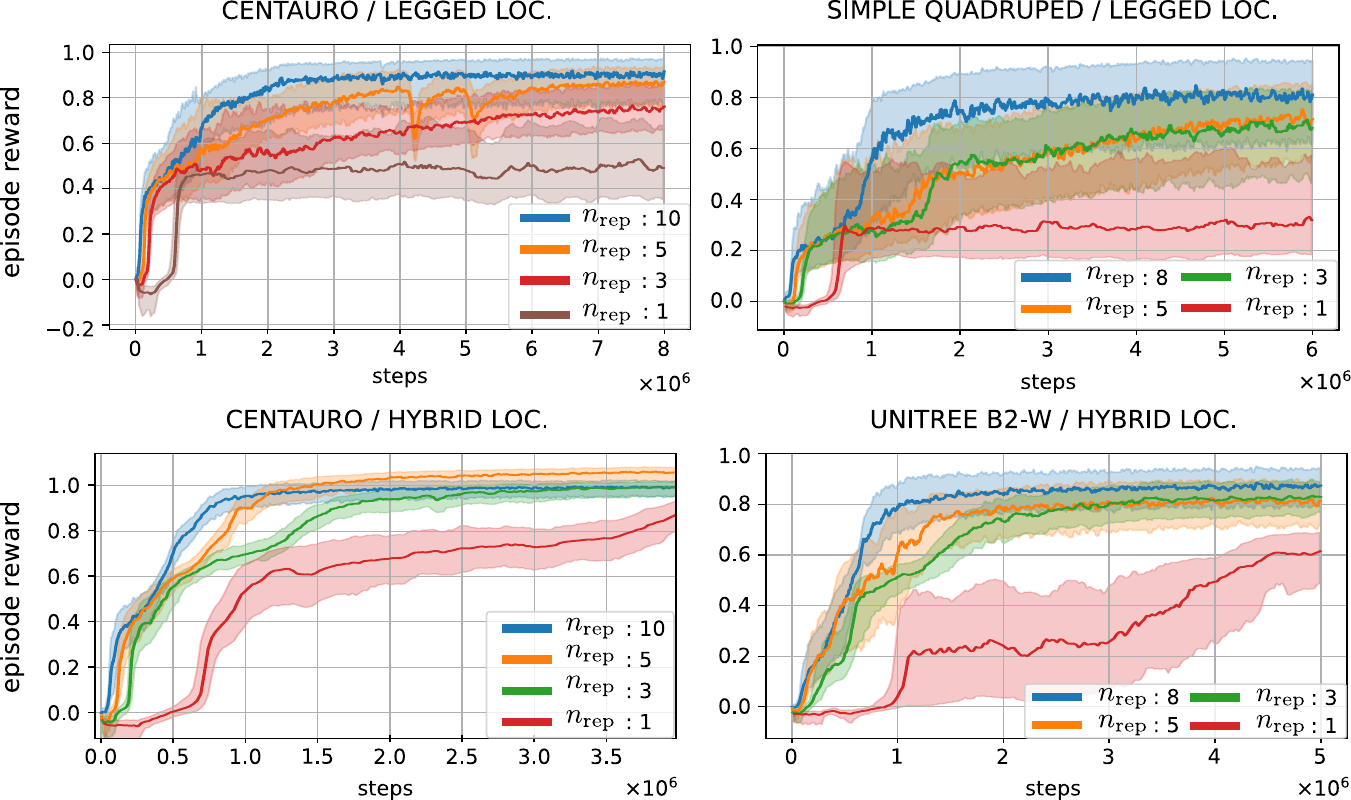}
	\caption{Influence of the action repeat parameter $n_{\mathrm{rep}}$ (see Sec.~\ref{subsec:mdp}) on the episode total reward (averaged across 800 environments) during training, shown across both legged and hybrid locomotion tasks. Shaded areas indicate the first and third quartiles of the distribution across environments.} 
	\label{fig:action_rep}
\end{figure}

One of the key hyperparameters of the proposed architecture is the policy frequency relative to the MPC, controlled by the parameter $n_{\mathrm{rep}}$ (see Sec.~\ref{sec:MDP}). We conduct an ablation study across multiple platforms to evaluate its impact on training performance, with results shown in Fig.~\ref{fig:action_rep}. 
For hybrid locomotion, the choice of $n_{\mathrm{rep}}$ has limited influence and generally does not hinder convergence. In contrast, for legged locomotion, $n_{\mathrm{rep}}$ plays a critical role: inappropriate values often lead to convergence to poor local minima, where the robot remains stationary and stepping behavior fails to emerge.
\subsection{Energy considerations}
Although energy comparisons across different robots and tasks can be misleading, a meaningful comparison can be drawn between legged and hybrid locomotion on Centauro, as the training setups are identical. As shown in Fig.~\ref{fig:sub_rewards}, the hybrid policy achieves substantially higher energy efficiency, with an average CoT at convergence of approximately $0.12$, compared to $0.35$ for legged locomotion. These results highlight the efficiency benefits of hybrid locomotion under identical task conditions.
\subsection{Domain Transfer Experiments}\label{subsec:transfer}

\revblock{In this section, we investigate the performance of the proposed architecture in transferring across simulators and to reality.}
In our hierarchical structure, the MPC partially decouples the learned policy from the environment, \revblock{substantially reducing the need for domain randomization}. However, real-world deployment can still introduce unmodeled noise into both the policy and MPC. We show how, despite this, our architecture remains effective and robust. 

Before moving into the real world, we evaluate policies for all the discussed platforms on a different simulator, Mujoco~\cite{sim:deepmind2021mujoco}, demonstrating \revblock{zero-shot} cross-simulator generalization. We also successfully deploy the full architecture on the real Centauro robot for both legged and hybrid locomotion tasks. Video frames of the sim-to-sim and sim-to-real experiments are shown in Fig.~\ref{fig:intro} and Fig.~\ref{fig:transfer_real_wheels}, while full video sequences are available in the supplementary material. 
Crucially, for the real-world deployment, we use policies trained with an MPC running in open-loop mode (see Sec.~\ref{subsec:closing_loop}).
Additionally, due to the lack of reliable onboard odometry, we estimate the robot's position directly from the MPC state, which proved to be sufficiently reliable.
\revblock{All computation, including policy inference and MPC evaluation, is executed in real time and onboard the robot.
}
\revblock{Across both sim-to-sim and sim-to-real experiments, the proposed architecture consistently exhibits non-periodic gait patterns, adaptive behaviors, and robust sustained operation in real-world legged and hybrid locomotion tasks.}
\begin{figure}[ht]
	\centering
	\vspace{0.3cm}
	\includegraphics[width=0.98\columnwidth]{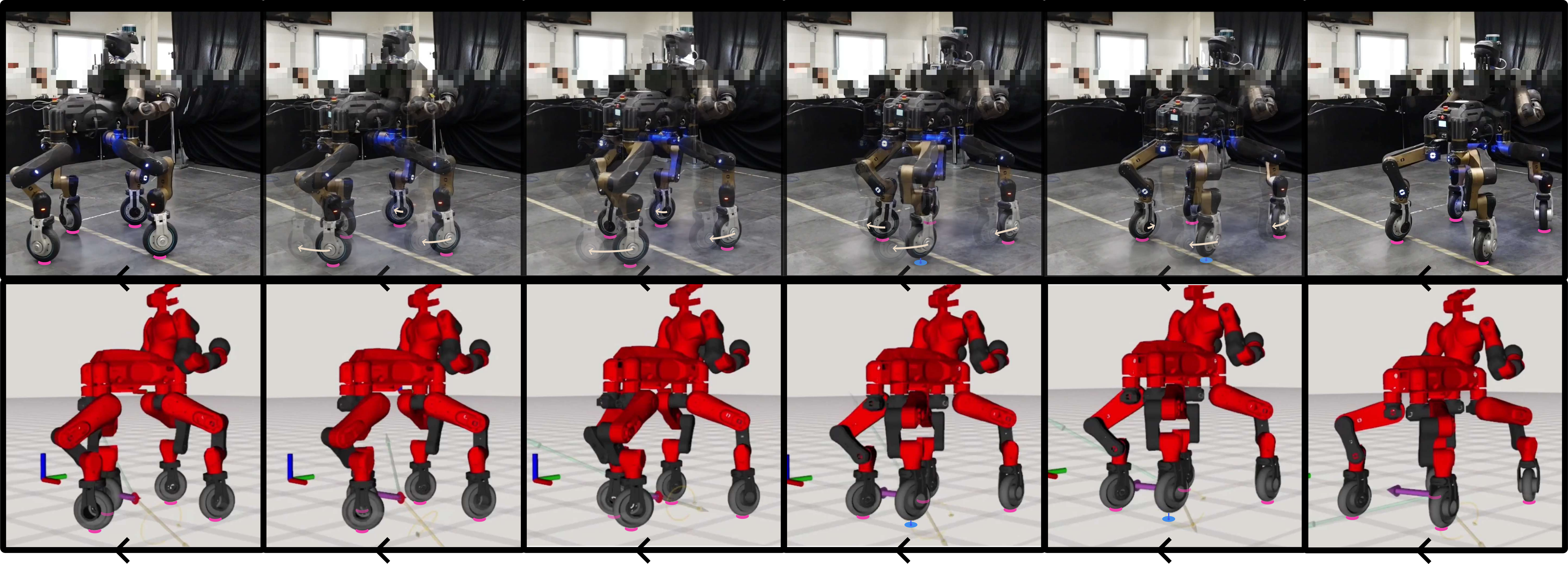}
	\caption{Zero-shot sim-to-real transfer of a hybrid locomotion policy on Centauro.} 
	\label{fig:transfer_real_wheels}
\end{figure}

\begin{figure}[ht]
	\centering
	\vspace{0.3cm}
	\includegraphics[width=0.98\columnwidth]{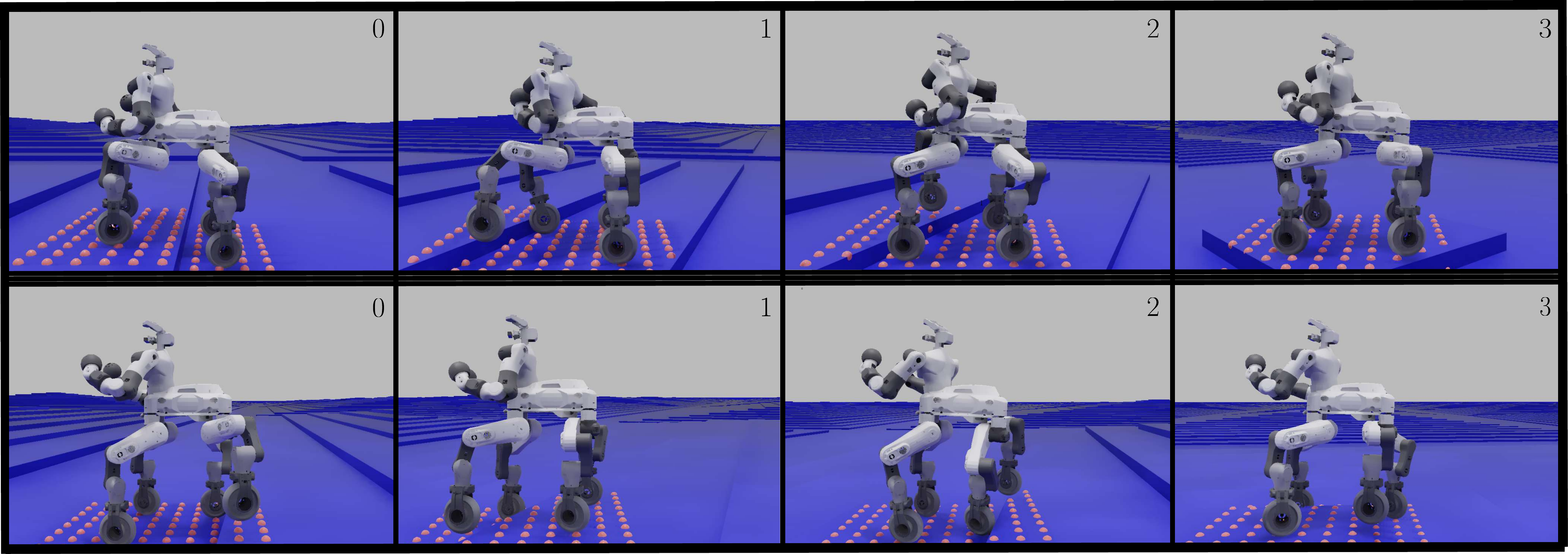}
	\caption{Evaluation of a perception-augmented hybrid locomotion policy for Centauro ascending a stepped pyramid. The policy also controls the flight phase parameters. Top row: ascent while facing forward. Bottom row: ascent while facing backward.} 
	\label{fig:non_flat}
\end{figure}

\revblock{
\subsection{Non-flat Terrains}\label{subsec:nonflat}
The proposed architecture can be extended to operate on non-flat terrains.
As a demonstration, we construct a hybrid locomotion task with Centauro using a full-body MPC, including upper-body control. We define a terrain consisting of randomly generated stepped pyramids, where each step is higher than the wheel radius, requiring stepping to ascend. We augment the agent’s observation with raw heightmap data and extend the MDP formulation of Sec.~\ref{subsec:mdp} to grant the policy full control over the flight phase parameters $H_{c}^{(j,\,\cdot)}$, $H_{l}^{(j,\,\cdot)}$, and $T_{\mathrm{fl}}^{(j,\,\cdot)}$ (see Sec.~\ref{subsec:horizon_control}).
Figure~\ref{fig:non_flat} shows the resulting policy successfully ascending a stepped pyramid while facing both forward and backward. Large steps emerge primarily during ascent, whereas during flat navigation the policy occasionally employs smaller steps to overcome non-holonomic constraints.
}

%% file: 06-future_work.tex
\section{Conclusions and future work}

We presented a contact-explicit hierarchical locomotion architecture combining a high-level RL policy with a low-level MPC controller for legged and hybrid locomotion. Our method eliminates the need for predefined gaits or demonstrations by learning acyclic contact schedules directly through trial and error. The system is easily adaptable to different robots, requires minimal tuning and reward engineering, and is capable of zero-shot sim-to-sim and sim-to-real transfer without domain randomization. These characteristics are enabled by a modular and scalable software framework that supports training with up to thousands of MPC instances in parallel.
Our simulated and real-world experiments demonstrate the emergence of non-periodic gaits and timing adaptations for both legged and hybrid locomotion. \revblock{Finally, we showed an example of how the proposed approach can be easily extended to handle unstructured environments by leveraging raw heightmap data and control over flight phase parameters.}

\revblock{Several directions for future work naturally arise from these results. While policies in this study are trained for specific MPC instances and robots, a natural next step is to investigate whether more general policies can be learned that transfer across controllers and platforms. In addition, future work will explore extensions of the RL agent’s action and observation space—building on the non-flat terrain example—with the aim of broadening applicability to more complex terrains and to other tasks, such as manipulation. For instance, enabling explicit control over contact landing locations may allow locomotion on terrains requiring precise foothold placement. Finally, a systematic evaluation of robustness to strong and persistent external disturbances remains an important direction for future investigation.}